\pdfoutput=1
\documentclass{article}
\usepackage{spconf,amsmath,graphicx}
\usepackage{hyperref}
\usepackage{color}

\title{SI-GAT: A METHOD BASED ON IMPROVED GRAPH ATTENTION NETWORK FOR SONAR IMAGE CLASSIFICATION}
\name{Can Lei\textsuperscript{1}, Huigang Wang\textsuperscript{1,2}, Juan Lei \textsuperscript{1} \thanks{Partially funding by National Science Foundation of China (62171368), Science, Technology and Innovation of Shenzhen Municipality(CYJ20190806150003606), the Fundamental Research Funds for the Central Universities(D5000220158).}}
\address{1. School of Marine Science and Technology, Northwestern Polytechnical University, Xi’an, China\\
	 2. Research $\& $ Development Institute of Northwestern Polytechnical University in Shenzhen, China
}
\begin{document}
	%\ninept
	%
	\maketitle
	\begin{abstract}
		The existing sonar image classification methods based on deep learning are often analyzed in Euclidean space, only considering the local image features. For this reason, this paper presents a sonar classification method based on improved Graph Attention Network (GAT), namely SI-GAT, which is applicable to multiple types imaging sonar. This method quantifies the correlation relationship between nodes based on the joint calculation of color proximity and spatial proximity that represent the sonar characteristics in non-Euclidean space, then the KNN (K-Nearest Neighbor) algorithm is used to determine the neighborhood range and adjacency matrix in the graph attention mechanism, which are jointly considered with the attention coefficient matrix to construct the key part of the SI-GAT. This SI-GAT is superior to several CNN (Convolutional Neural Network) methods based on Euclidean space through validation of real data.
	\end{abstract}
	\section{Introduction}
	\label{sec:intro}
	
	Underwater target classification is one of the most challenging applications for sonar, which can be used for underwater search and rescue \cite{A1}, submarine pipeline detection \cite{A2}, anti-mine and frogman \cite{A3}. Sonar target classification is mainly presented in two forms: instantaneous acoustic signal detection and regional imaging detection \cite{A4}, among which two kinds of imaging sonar, multi-beam and side-scan sonar (SSS), are more widely used.
	
	With the great success of deep learning in optical image recognition, this method has also been introduced into the field of sonar image processing \cite{A5,A6}. NATO Ocean Research and Experiment Center firstly applied CNN to underwater sonar image classification by image fusion technology combined with deep learning algorithm in 2016 \cite{A7}. Subsequently, many researches have made improvements on this basis \cite{A8,A9}. \cite{A10} proposed a new underwater sonar image depth learning model based on adaptive weight convolution neural network (AW-CNN), which can automatically extract image features through the internal network structure. \cite{A11} proposed a lightweight recursive transitive adaptive learning (RTAL) algorithm, applies the lightweight recursive transitive learning (RTL) to SSS image recognition.
	
	The above methods with CNN mode are analyzed under Euclidean space, where sonar image is participated into regular grid region without special correlation \cite{A12}. However, Graph neural networks (GNN) based on non-Euclidean space will combine multiple features to enhance the correlation between different regions and extract richer features \cite{A13,A14}. \cite{A15} proposed a weighted feature fusion of convolution neural network and graphical attention network (WFCG) for hyperspectral imaging (HSI) classification by utilizing the features of GAT based on superpixels and CNN based on pixels. \cite{A16} presented a graphical attention network-driven robust representation learning method for multilabel images (RRL-GAT) to reduce noise and spurious connections between objects. \cite{A17} presented a graphical attention-based multimodal HAR method, Multi-GAT, to learn complementary multimodal features hierarchically for feature interaction. However, to the best of my knowledge, no GNN (Graph Neural Network) has been found to be used in the underwater sonar imaging.
	
	Based on the above considerations, this paper proposes a method based on improved graph attention network for sonar imaging, namely SI-GAT, which is applicable to both side-scan sonar images and multi-beam sonar images. 

	\section{The SI-GAT method}
	In the original GAT model\cite{A18}, the neighborhood space is selected by the rule of distance, and the neighborhood nodes only contain first-order neighbor, which is inappropriate for sonar image \cite{A19}. Therefore, we propose SI-GAT method for the application scenario of sonar images on this basis. In SI-GAT, the selection rule of neighbor nodes are defined and explained in detail according to the particularity of sonar images. In addition, in order to reduce computational effort and eliminate excessive noise caused by redundant relationships between nodes, the method considers both adjacency matrix and attention factor matrix adjusted by the KNN algorithm. The specific process is shown in Fig. 1 below.
	\begin{figure}[ht]
		\centering
		\includegraphics[width=1\linewidth]{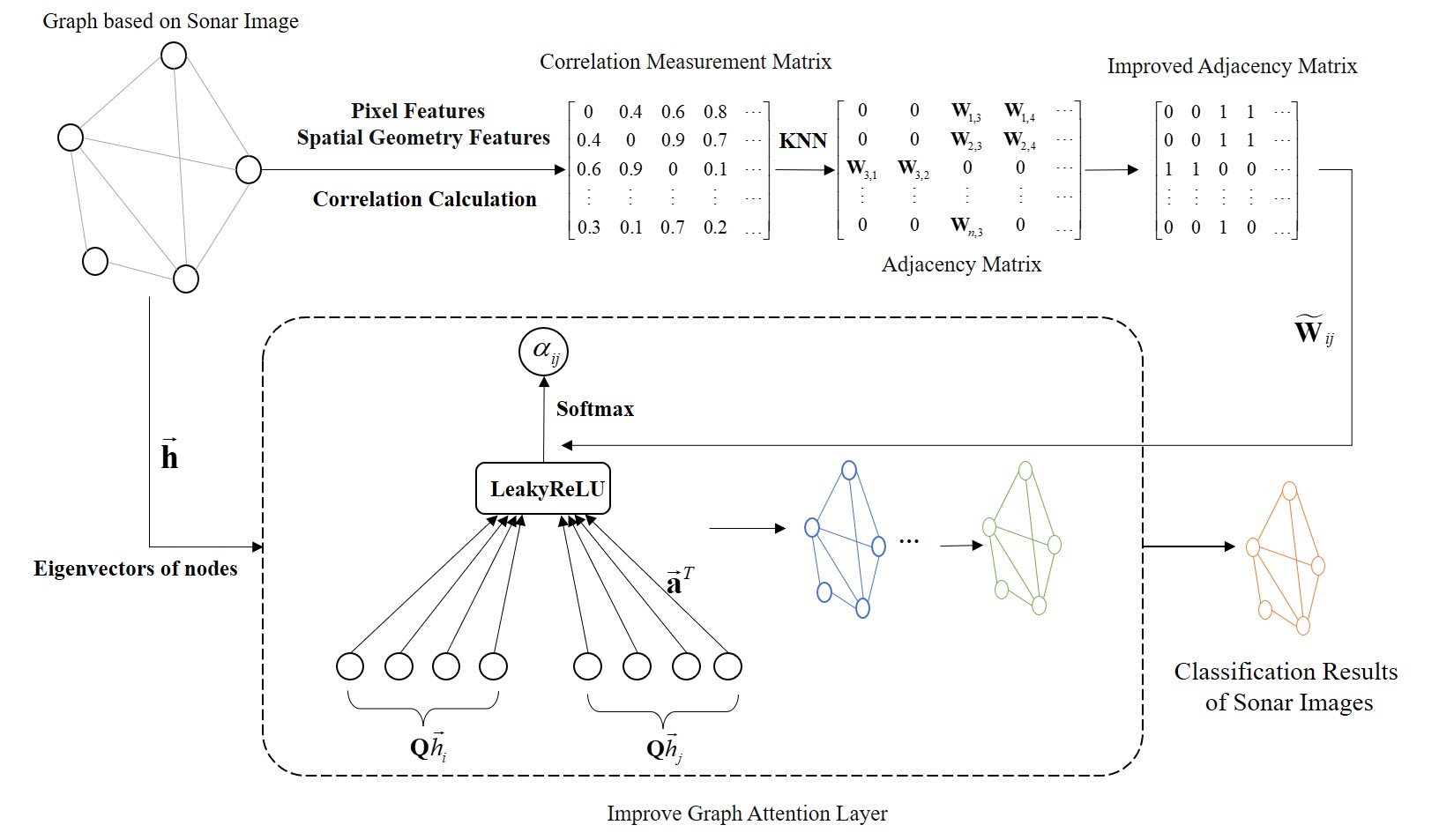}
		\caption{The block diagram of the SI-GAT method.}
		\label{fig1}
	\end{figure}	
	\subsection{Correlation measurement matrix}
	The correlation measurement matrix in GAT network stores the connection relationship between all pairwise nodes, which is the edge weight \cite{A20}. According to the special imaging properties of sonar image, where the imaging principle of side-scan sonar is shown in Fig. 2, a target-related acoustic shadow area will be formed near target area, which leads to the inclusion of three regions in one sonar image: the target region, the acoustic shadow region and the reverberation background.  Therefore, the nodes formed by sonar image not only contain the pixel features but also their spatial position features, where the pixel features reflect the edge information of the sonar target and the spatial location features reflect the correlation between the bright and shaded regions of the target, so the matrix generated by the original GAT that only contains 0 (no connection between nodes) and 1 (connection between nodes) is not applicable to sonar images \cite{A19}. In order to ensure the correctness of the subsequent selection of neighborhood nodes, we propose a new method to compute the correlation measurement matrix for sonar images.
	\begin{figure}[ht]
		\centering
 		\includegraphics[width=0.8\linewidth]{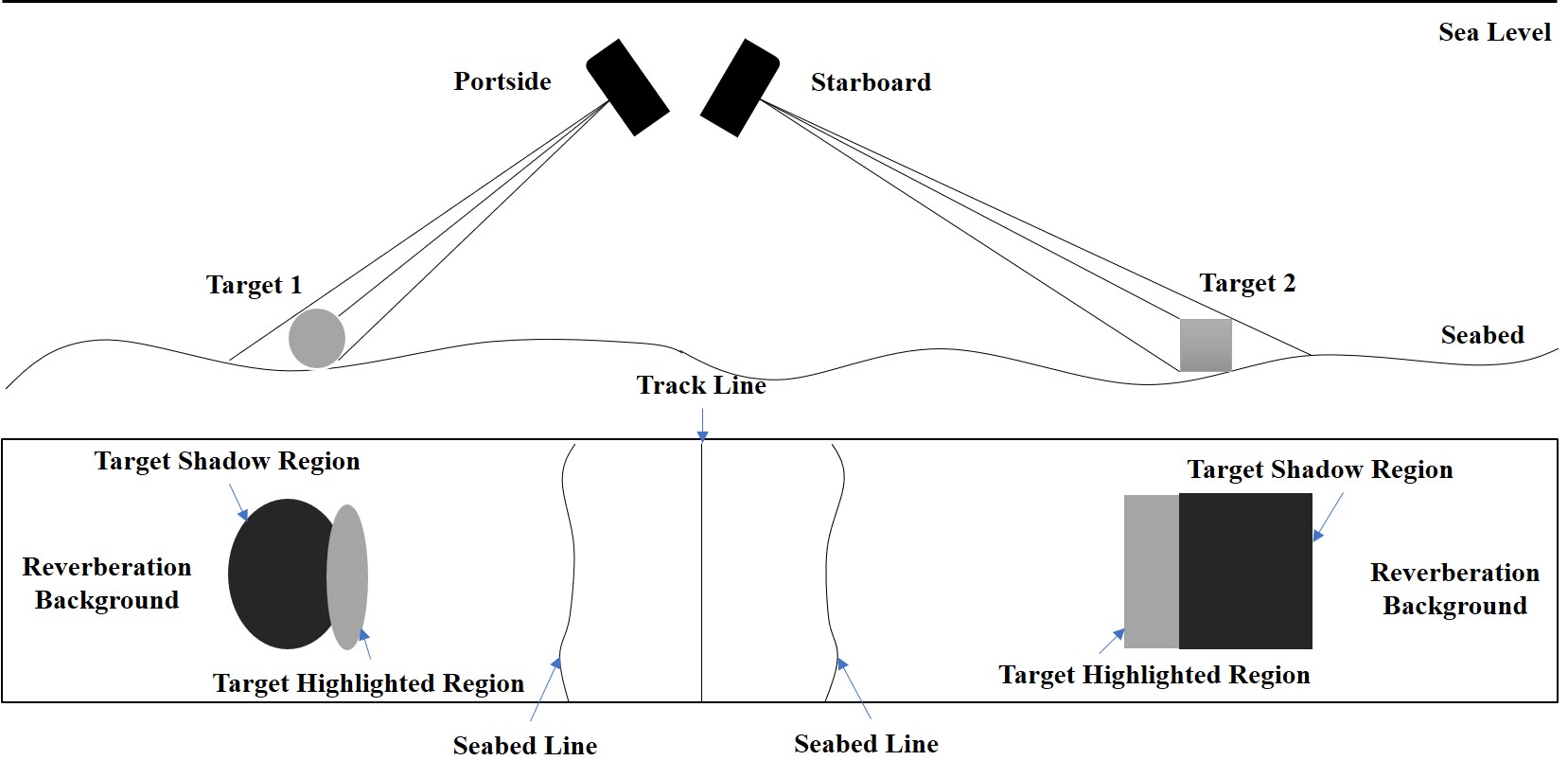}
		\caption{The imaging principle of side-scan sonar image.}
		\label{fig2}
	\end{figure}
	
	Since the range of the horizontal and vertical position coordinates of the sonar image are $\left[ {0,{x_{max}}} \right],\left[ {0,{y_{max}}} \right]$, and the range of the pixel values is $[0,1]$, the position coordinates need to be normalized firstly, which is calculated as:$({\widehat x_i},{\widehat y_i}) = ({x_i}/{x_{max}},{y_i}/{y_{max}})$. Thus, the values of each element ${{\bf{W}}_{i,j}}$ in the correlation measurement matrix are calculated as follows:
	\begin{equation}
		{{\bf{W}}_{i,j}} = \gamma {f_{coord}}({\widehat x_i},{\widehat y_i}) + (1 - \gamma ){f_{pix}}({\widehat x_i},{\widehat y_i})
	\end{equation}
    where $\gamma $ represents the relative important factor between the pixel information and the spatial location information, ${f_{coord}}({\widehat x_i},{\widehat y_i})$ represents the spatial position correlation degree, and ${f_{pix}}({\widehat x_i},{\widehat y_i})$ represents the pixel correlation degree. 
    
    For spatial location feature correlation calculation, the basic rule is that the farther the node from the reference pixel is, the weaker the correlation is, and vice versa. Therefore, the spatial location association of nodes can be well represented by exponential function ${e^{ - x}}$. Specific spatial location correlation is calculated as follows:
    \begin{equation}
    	{f_{coord}}({\widehat x_i},{\widehat y_i}) = \exp \left( { - \frac{{{{\left\| {({{\widehat x}_i},{{\widehat y}_i}) - ({{\widehat x}_j},{{\widehat y}_j})} \right\|}^2}}}{{\delta _x^2}}} \right)
    \end{equation}
    where ${({{\widehat x}_i},{{\widehat y}_i})}$ represents the normalized location information of the $ith$ node and ${\delta _x}$ represents the average value of the spatial location distance between the $ith$ node and other nodes.
	
	For pixel feature correlation calculation, in order to represent the boundaries of the three regions in the mentioned sonar image above when computing the correlation using the pixel information, the ${e^{ - x}}$ function is no longer applicable, but the Sigmoid function applies exactly to this rule \cite{A21}. The specific calculation of pixel correlation is as follows:
	\begin{equation}
		{f_{pix}}({\widehat x_i},{\widehat y_i}) = \frac{1}{{1 + \exp \left( { - {{\left( {\frac{{\left\| {f({{\widehat x}_i},{{\widehat y}_i}) - f({{\widehat x}_j},{{\widehat y}_j})} \right\|}}{{\delta _f^2}}} \right)}^2}} \right)}}
	\end{equation}
	where $f({\hat x_i},{\hat y_i})$ represents the pixel information of $ith$ node and ${\delta _f}$ represents the average pixel distance between the $ith$ node and other nodes. Based on the above Eq. (1-3), each element value in the final correlation measurement matrix is calculated as follows:
	\begin{small}
	\begin{equation}
		\begin{split}
			{{\bf{W}}_{i,j}} & = \gamma \exp \left( { - \frac{{{{\left\| {({{\widehat x}_i},{{\widehat y}_i}) - ({{\widehat x}_j},{{\widehat y}_j})} \right\|}^2}}}{{\delta _x^2}}} \right)\\
			& + (1 - \gamma )\frac{1}{{1 + \exp \left( { - {{\left( {\frac{{\left\| {f({{\widehat x}_i},{{\widehat y}_i}) - f({{\widehat x}_j},{{\widehat y}_j})} \right\|}}{{\delta _f^2}}} \right)}^2}} \right)}}
		\end{split}
	\end{equation}
    \end{small}
	\subsection{KNN model}
	Once the correlation measurement matrix of the graph structure is obtained, the nodes in the neighborhood need to be selected. In the original GAT, first-order neighbor nodes are selected from a set of connected edges, but this method is not appropriate in the field of sonar image because the pixel features in sonar images cannot be utilized in the above rule. Therefore, in order to make full use of various features and eliminate computational redundancy, the SI-GAT method uses KNN algorithm \cite{A22} to find closely related edges that need to be preserved and low correlation edges that need to be disconnected.
	
	This method first sorts the connectivity (edge weight value) from each node to the rest of the nodes from strong to weak:
	\begin{equation}
		Arr{a_i} = \mathop {Rank}\limits_{e \in {N_i}} \left\{ {{{\bf{W}}_{i,1}},{{\bf{W}}_{i,2}}{{\bf{W}}_{i,3}} \cdots {{\bf{W}}_{i,e}} \cdots {{\bf{W}}_{i,n}}} \right\}
	\end{equation}
	where $n$ is the total number of nodes. Then we choose an appropriate $k$ value, and top $k$ edges with strong correlation are selected and retained according to the sorting result:
	\begin{equation}
		{N_i} = \left\{ {Arr{a_{i,1}},Arr{a_{i,2}},Arr{a_{i,3}} \cdots Arr{a_{i,k}}} \right\}
	\end{equation}
	
	After obtaining $k$ edges, the edge with weak connectivity needs to be deleted for each node, that is, the remaining edge connections are broken, and the weight of the edge (the corresponding element value in the correlation measurement matrix) is reassigned to 0. Finally, according to the KNN module, the adjacency matrix ${\bf{A}}$ and the updated edge weights ${{\bf{W'}}_{i,j}}$ are obtained as follows:
	
	\begin{scriptsize}
	\begin{equation}
		{{\bf{W'}}_{i,j}} = \left\{ \begin{array}{l}
			\gamma \exp \left( { - \frac{{{{\left\| {({{\widehat x}_i},{{\widehat y}_i}) - ({{\widehat x}_j},{{\widehat y}_j})} \right\|}^2}}}{{\tilde \delta _x^2}}} \right)\\
			+ (1 - \gamma )\frac{1}{{1 + \exp \left( { - {{\left( {\frac{{\left\| {f({{\widehat x}_i},{{\widehat y}_i}) - f({{\widehat x}_j},{{\widehat y}_j})} \right\|}}{{\tilde \delta _f^2}}} \right)}^2}} \right)}},j \in {N_i}\\
			0,j \notin {N_i}
		\end{array} \right.
	\end{equation}
    \end{scriptsize}
    \begin{scriptsize}
    \begin{equation}
    		{\bf{A}} =  KNN {\bf{(W')}} = \left[ {\begin{array}{*{20}{c}}
    				{\rm{0}}&{\rm{0}}&{{{\bf{W'}}_{1,3}}}&{{{\bf{W'}}_{1,4}}}& \cdots \\
    				{\rm{0}}&{\rm{0}}&{{{\bf{W'}}_{2,3}}}&{{{\bf{W'}}_{2,4}}}& \cdots \\
    				{{{\bf{W'}}_{3,1}}}&{{{\bf{W'}}_{3,2}}}&{{{\bf{W'}}_{3,3}}}&{\rm{0}}& \cdots \\
    				\vdots & \vdots & \vdots & \vdots & \cdots \\
    				{\rm{0}}&{\rm{0}}&{{{\bf{W'}}_{n,3}}}&{\rm{0}}& \ldots 
    		\end{array}} \right]
    \end{equation}
    \end{scriptsize}where ${N_i}$ is the neighborhood range of the $ith$ node, ${\tilde \delta _x^2}$ is the average spatial location distance between the $ith$ node and its nearest $k$ nodes, and ${\tilde \delta _f^2}$ represents the average pixel distance between the $ith$ node and its nearest $k$ nodes.

	\subsection{Improved Graph Attention Layer}
	The input to the graph attention layer is the feature combination for each node, denoted as: ${\bf{h}} = \left\{ {{{\vec h}_1},{{\vec h}_2}, \ldots ,{{\vec h}_M}} \right\}$, ${\vec h_i} \in {{\mathop{\rm R}\nolimits} ^F}$, where $M$ is the number of nodes, ${h_i}$ is the eigenvector of the $ith$ node, and $F$ is the number of features for each node. Node are superpixels obtained from sonar images, and the features are the pixel features and spatial location features obtained from correlation measurement matrix. After the input passes through the attention layer, a new node feature combination ${{\bf{h}}^\prime } = \left\{ {\vec h_1^\prime ,\vec h_2^\prime , \ldots ,\vec h_M^\prime } \right\},\vec h_i^\prime  \in {{\mathop{\rm R}\nolimits} ^{{F^\prime }}}$ is obtained from the output. The shared linear transformation of each node between input and output is defined as a parameterized weight matrix ${\bf{Q}} \in {{\mathop{\rm R}\nolimits} ^{{F^\prime } \times F}}$ \cite{A18}. Since the attention layer incorporates the attention mechanism and assigns different coefficient weights between the current node and its neighbor nodes, the input and output of the whole attention layer can be represented as follows:
	\begin{equation}
		\vec h_i^\prime  = \sigma \left( {\sum\limits_{j \in {{\cal N}_i}} {{\alpha _{ij}}} {\bf{Q}}{{\vec h}_j}} \right)
	\end{equation}
	where ${\alpha _{ij}}$ is the attention coefficient (coefficient weight) between the $ith$ node and the $jth$ node, and $\sigma $ is the nonlinear activation function. The weight matrix ${\bf{Q}}$ is obtained through back propagation, $\sigma $ is a predefined activation function, then only the attention coefficient $\alpha $ is unknown in the above equation \cite{A19}, which needs to be calculated by the following equation: 
	
	\begin{footnotesize}
	\begin{equation}
		{\alpha _{ij}} = \frac{{\exp \left( {{\mathop{\rm LeakyReLU}\nolimits} \left( {{{\overrightarrow {\bf{a}} }^T}\left[ {{\bf{Q}}{{\vec h}_i}\parallel {\bf{Q}}{{\vec h}_j}} \right]} \right)} \right)}}{{\sum\limits_{k \in {{\cal N}_i}} {\exp } \left( {{\mathop{\rm LeakyReLU}\nolimits} \left( {{{\overrightarrow {\bf{a}} }^T}\left[ {{\bf{Q}}{{\vec h}_i}\parallel {\bf{Q}}{{\vec h}_k}} \right]} \right)} \right)}}
 	\end{equation}
    \end{footnotesize}
    where $\mathop a\limits^ \to   \in {{\mathop{\rm R}\nolimits} ^{2F^\prime}}$ is the parameterization of the self-attention mechanism, $\parallel $ is the splicing operation, and $k$ is the number of neighbors selected. In the above expression of attention coefficient, there are two unknown points: the number of nodes contained in the neighborhood space and the specific location of the neighborhood selection. In this method, the results obtained by KNN module are used to solve the above two problems. In addition, in order to simplify the calculation and enhance the difference, the weight ${{\bf{W'}}_{i,j}}$ of top $k$ edges with strong connectivity is reassigned to 1, and the weight of the remaining edges is assigned to 0, and then the updated adjacency matrix $\widetilde {\bf{A}}$ and edge weight ${{{\widetilde {\bf{W}}}_{i,j}}}$ are obtained as follows: 
    \begin{small}
    \begin{equation}
    	{\widetilde {\bf{W}}_{i,j}} = \left\{ \begin{array}{l}
    		{\rm{1}},j \in {N_i}\\
    		0,j \notin {N_i}
    	\end{array} \right.,\widetilde {\bf{A}} = \left[ {\begin{array}{*{20}{c}}
    			0&0&{\rm{1}}&{\rm{1}}& \cdots \\
    			0&0&{\rm{1}}&{\rm{1}}& \cdots \\
    			{\rm{1}}&{\rm{1}}&{\rm{1}}&0& \cdots \\
    			\vdots & \vdots & \vdots & \vdots & \cdots \\
    			0&0&{\rm{1}}&0& \ldots 
    	\end{array}} \right]
    \end{equation}
    \end{small}

    Before SoftMax normalization, the matrix $\widetilde {\bf{A}}$ is multiplied with the attention coefficient matrix to determine the selection of neighborhood nodes, so as to obtain the final expression of the attention mechanism coefficient as follows:
    
    \begin{footnotesize}
    	\begin{equation}
    		{\alpha _{ij}} = \frac{{\exp \left( {{{\widetilde {\bf{W}}}_{i,j}}\left( {{\mathop{\rm LeakyReLU}\nolimits} \left( {{{\overrightarrow {\bf{a}} }^T}\left[ {{\bf{Q}}{{\vec h}_i}\parallel {\bf{Q}}{{\vec h}_j}} \right]} \right)} \right)} \right)}}{{\sum\limits_{k \in {{\cal N}_i}} {\exp } \left( {{{\widetilde {\bf{W}}}_{i,k}}\left( {{\mathop{\rm LeakyReLU}\nolimits} \left( {{{\overrightarrow {\bf{a}} }^T}\left[ {{\bf{Q}}{{\vec h}_i}\parallel {\bf{Q}}{{\vec h}_k}} \right]} \right)} \right)} \right)}}
    	\end{equation}
    \end{footnotesize}
    
    The attention coefficients between different nodes are calculated from Eq. (12), and the coefficients are taken into the calculation formula between the input and output of the convolutional layer to obtain the final output features of each node. When the construction of the improved graph attention layer is completed, the appropriate number $L$ of attention layers is selected to construct the whole network model \cite{A18}.

	\section{Experiment}
	
	\subsection{Data Description}
	
	The multi-beam sonar image dataset collected from Pengcheng Laboratory \footnote{https://challenge.datacastle.cn/v3/cmptDetail.html?id=680} contains five types of floating target: human body, tire, column, sphere and cube. A total of 803 chart structures are created, where there are 300 sphere images, 204 cube images, 85 column images, 103 dummy images, and 111 tire images. The total of 695 side-scan sonar images are sorted out, among which the sonar images of plane and ship are collected from the dataset publicly shared on github by Huo Guanying's team \footnote{https://github.com/huoguanying/SeabedObjects-Ship-and-Airplane-dataset.git}, and the data of mines and human bodies are chosen from the gallery display list of various sonar software companies (such as Tritech, Edgetech, etc.). All data are divided into training set, validation set and test set according to the proportion of 0.7:0.1:0.2.
	
	\subsection{Experimental Settings and Model Training}
	
	This method uses DGL (Deep Graph Library) as the graph neural network framework to complete the model construction under the PyTorch framework, and the CPU is Intel ® Xeon ® Silver 4110 CPU @ 2.10GHz, GPU is NVIDIA GeForce RTX 3080. 
	
	The total parameter number of the SI-GAT is 111078, and the number of graph attention layers  is set to 4 with 9 hidden layers. The dimension of the output feature vector is set to 152, the number of attention mechanisms is set to 8, and the node aggregation method is set to the average value. 
	
	In the training process, the side-scan sonar image size is patched to 200x200 pixels, the looking forward sonar image size is limited to 200x600. The epochs length is set to 250, and the batch size is set to 4 with the initial learning rate 0.001 and the learning rate attenuation coefficient with 0.5.
	
	\subsection{Experimental Results}
	
	To validate the effectiveness of the proposed SI-GAT for high-precision classification of sonar images, we selected three graph neural networks based on non-Euclidean space: GAT, GatedGCN (Gated Graph Convolutional Network), GCN (Graph Convolutional Network), and three CNN networks based on Euclidean space: VGG16, Mobilenet, Resnet50. The parameters number of six networks are shown in Table 1. To be fair, all networks are trained from scratch, and the same data sets are sent to six networks. 
	
	\begin{table}[h]\small
		\begin{center}
			\renewcommand{\arraystretch}{1.2}
			\caption{Table of the total parameter numbers for the six networks.}
			\setlength{\tabcolsep}{1mm}{
				\begin{tabular}{| c | c | c |c |c |c |c |}
					\hline
					SI-GAT & GAT & GatedGCN & GCN & VGG16 & Mobilenet & Resnet50 \\
					\hline
					0.11M & 0.11M & 0.10M & 0.10M & 138M & 5.34M & 23.56M\\
					\hline		
			\end{tabular}}
		\end{center}
	\end{table}	

	As shown in Fig. 3, SI-GAT model is optimal from the perspective of training curve and recognition rate. In the training curve, all models of graph neural network converge faster than those with CNN structure due to smaller parameters and non-Euclidean space. In the recognition stage, the SI-GAT has 3.25\% higher recognition rate than Resnet50, which is the best one in all CNN networks.
	\begin{figure}[htbp]
		\begin{minipage}[t]{0.5\linewidth}
			\centering
			\includegraphics[width=\textwidth]{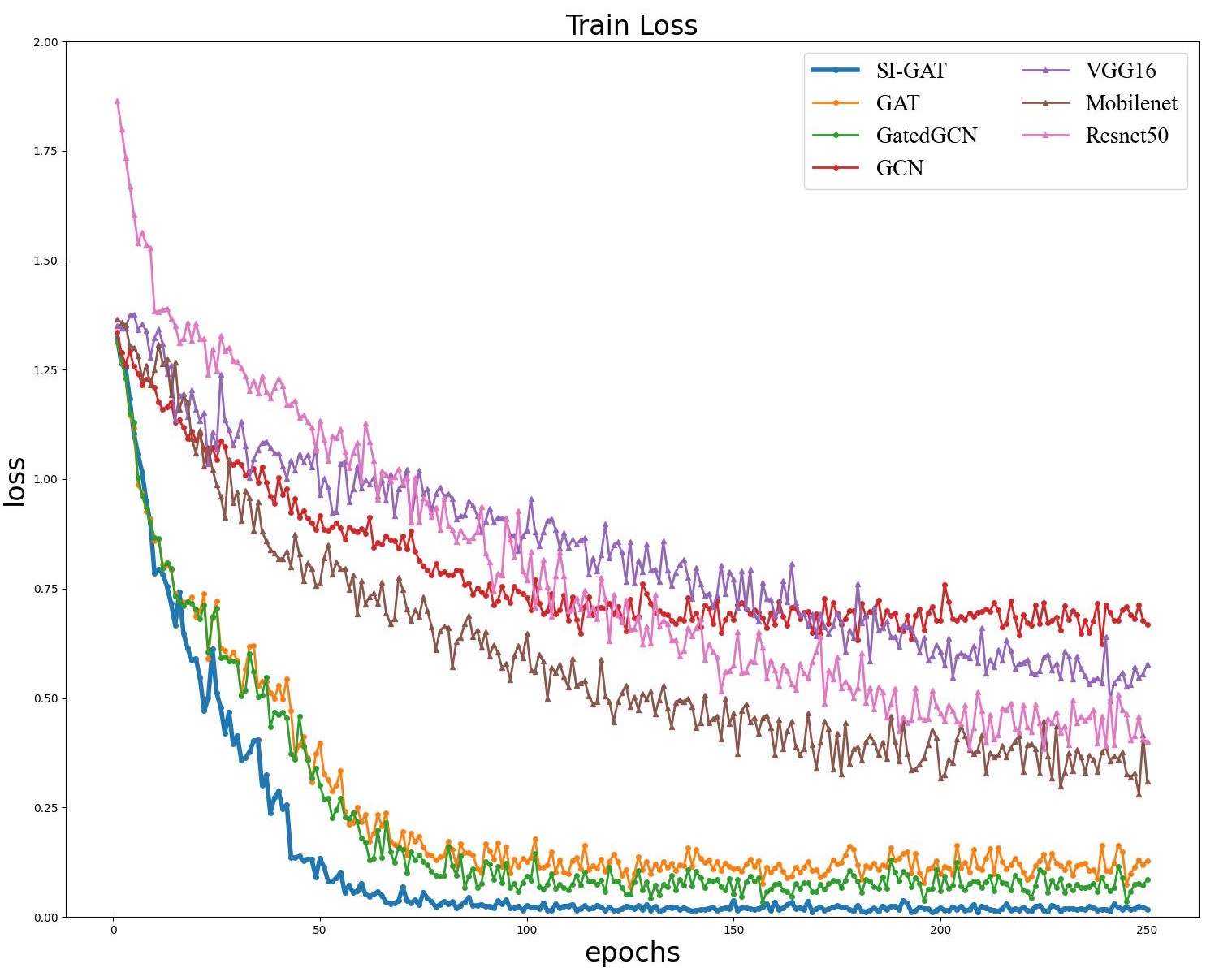}
			\centerline{(a)}
		\end{minipage}%
		\begin{minipage}[t]{0.45\linewidth}
			\centering
			\includegraphics[width=\textwidth]{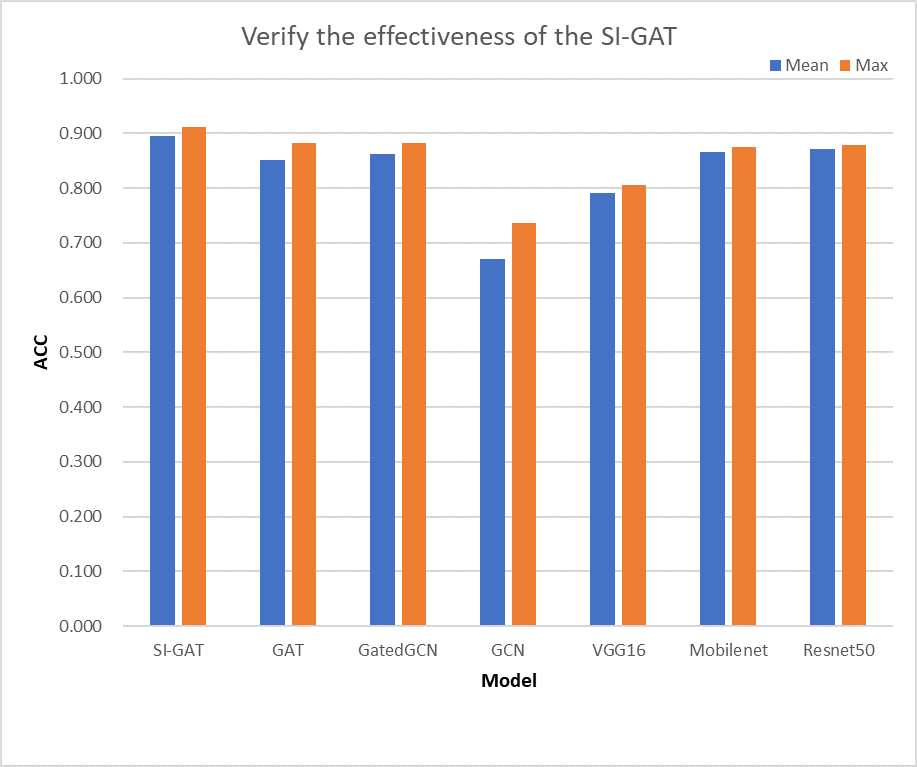}
			\centerline{(b)}
		\end{minipage}
		\caption{(a)The train loss for all models, from left to right: SI-GAT, GAT, GatedGCN, GCN, VGG16, Mobilenet and Resnet50. (b)The classification accuracy for all models.}
	\end{figure}
%	\begin{figure}[ht]
%		\centering
%		\includegraphics[width=0.8\linewidth]{train_loss.jpg}
%		\caption{The train loss for all models, from left to right: SI-GAT, GAT, GatedGCN, GCN, VGG16, Mobilenet and Resnet50.}
%		\label{fig3}
%	\end{figure}
%
%	\begin{figure}[ht]
%		\centering
%		\includegraphics[width=0.8\linewidth]{compare.png}
%		\caption{The classification accuracy for all models, from left to right: SI-GAT, GAT, GatedGCN, GCN, VGG16, Mobilenet and Resnet50.}
%		\label{fig4}
%	\end{figure}
	
	\section{Conclusion}
	\label{sec:illust}
	This paper presents a SI-GAT method based on improved graphical attention network, which is suitable for both side-scan sonar image and multi-beam sonar image classification. This method extracts more detailed and global features by jointly considering the pixel and position information of sonar images to achieve higher accuracy classification. Another advantage of the SI-GAT is fast convergence speed in the case of few samples because of small number of parameters.

	% Below is an example of how to insert images. Delete the ``\vspace'' line,
	% uncomment the preceding line ``\centerline...'' and replace ``imageX.ps''
	% with a suitable PostScript file name.
	% -------------------------------------------------------------------------
%	\begin{figure}[htb]
%		
%		\begin{minipage}[b]{1.0\linewidth}
%			\centering
%			\centerline{\includegraphics[width=8.5cm]{image1}}
%			%  \vspace{2.0cm}
%			\centerline{(a) Result 1}\medskip
%		\end{minipage}
%		%
%		\begin{minipage}[b]{.48\linewidth}
%			\centering
%			\centerline{\includegraphics[width=4.0cm]{image3}}
%			%  \vspace{1.5cm}
%			\centerline{(b) Results 3}\medskip
%		\end{minipage}
%		\hfill
%		\begin{minipage}[b]{0.48\linewidth}
%			\centering
%			\centerline{\includegraphics[width=4.0cm]{image4}}
%			%  \vspace{1.5cm}
%			\centerline{(c) Result 4}\medskip
%		\end{minipage}
%		%
%		\caption{Example of placing a figure with experimental results.}
%		\label{fig:res}
%		%
%	\end{figure}
%	
	
	%\vfill
	%\pagebreak

	\bibliographystyle{unsrt}
	\bibliography{referen}

\begin{thebibliography}{10}

\bibitem{A1}
Kazuoki Kuramoto, Akira Asada, and Kazuhiro Hantani.
\newblock Development of the underwater search supporting technique using the
  high-resolution imaging sonar.
\newblock In {\em OCEANS 2015-MTS/IEEE Washington}, pages 1--6. IEEE, 2015.

\bibitem{A2}
Shaobo Li, Jianhu Zhao, Hongmei Zhang, and YouPeng Zhang.
\newblock Automatic detection of pipelines from sub-bottom profiler sonar
  images.
\newblock {\em IEEE Journal of Oceanic Engineering}, 47(2):417--432, 2021.

\bibitem{A3}
Narc{\'\i}s Palomeras, Thomas Furfaro, David~P Williams, Marc Carreras, and
  Samantha Dugelay.
\newblock Automatic target recognition for mine countermeasure missions using
  forward-looking sonar data.
\newblock {\em IEEE Journal of Oceanic Engineering}, 47(1):141--161, 2021.

\bibitem{A4}
Lei Wang.
\newblock {\em Research on side-scan sonar image segmentation algorithm}.
\newblock PhD thesis, Harbin: Harbin Engineering University, 2013.

\bibitem{A5}
Xingmei Wang, Jia Jiao, Jingwei Yin, Wensheng Zhao, Xiao Han, and Boxuan Sun.
\newblock Underwater sonar image classification using adaptive weights
  convolutional neural network.
\newblock {\em Applied Acoustics}, 146:145--154, 2019.

\bibitem{A6}
Xu~Cao, Roberto Togneri, Xiaomin Zhang, and Yang Yu.
\newblock Convolutional neural network with second-order pooling for underwater
  target classification.
\newblock {\em IEEE Sensors Journal}, 19(8):3058--3066, 2018.

\bibitem{A7}
David~P Williams and Samantha Dugelay.
\newblock Multi-view sas image classification using deep learning.
\newblock In {\em OCEANS 2016 MTS/IEEE Monterey}, pages 1--9. IEEE, 2016.

\bibitem{A8}
Dylan Einsidler, Manhar Dhanak, and Pierre-Philippe Beaujean.
\newblock A deep learning approach to target recognition in side-scan sonar
  imagery.
\newblock In {\em OCEANS 2018 MTS/IEEE Charleston}, pages 1--4. IEEE, 2018.

\bibitem{A9}
Youngmin Choo, Keunhwa Lee, Wooyoung Hong, Sung-Hoon Byun, and Haesang Yang.
\newblock Active underwater target detection using a shallow neural network
  with spectrogram-based temporal variation features.
\newblock {\em IEEE Journal of Oceanic Engineering}, 2022.

\bibitem{A10}
Xingmei Wang, Jia Jiao, Jingwei Yin, Wensheng Zhao, Xiao Han, and Boxuan Sun.
\newblock Underwater sonar image classification using adaptive weights
  convolutional neural network.
\newblock {\em Applied Acoustics}, 146:145--154, 2019.

\bibitem{A11}
Fei Yu, Bo~He, and Ji-Xin Liu.
\newblock Underwater targets recognition based on multiple auvs cooperative via
  recurrent transfer-adaptive learning (rtal).
\newblock {\em IEEE Transactions on Vehicular Technology}, 2022.

\bibitem{A12}
Shengxi Jiao, Chunyu Zhao, and Ye~Xin.
\newblock Research on convolutional neural network model for sonar image
  segmentation.
\newblock In {\em MATEC Web of Conferences}, volume 220, page 10004. EDP
  Sciences, 2018.

\bibitem{A13}
Zonghan Wu, Shirui Pan, Fengwen Chen, Guodong Long, Chengqi Zhang, and S~Yu
  Philip.
\newblock A comprehensive survey on graph neural networks.
\newblock {\em IEEE transactions on neural networks and learning systems},
  32(1):4--24, 2020.

\bibitem{A14}
Vijay~Prakash Dwivedi, Anh~Tuan Luu, Thomas Laurent, Yoshua Bengio, and Xavier
  Bresson.
\newblock Graph neural networks with learnable structural and positional
  representations.
\newblock {\em arXiv preprint arXiv:2110.07875}, 2021.

\bibitem{A15}
Yanni Dong, Quanwei Liu, Bo~Du, and Liangpei Zhang.
\newblock Weighted feature fusion of convolutional neural network and graph
  attention network for hyperspectral image classification.
\newblock {\em IEEE Transactions on Image Processing}, 31:1559--1572, 2022.

\bibitem{A16}
Bin Hu, Kehua Guo, Xiaokang Wang, Jian Zhang, and Di~Zhou.
\newblock Rrl-gat: Graph attention network-driven multi-label image robust
  representation learning.
\newblock {\em IEEE Internet of Things Journal}, 2021.

\bibitem{A17}
Md~Mofijul Islam and Tariq Iqbal.
\newblock Multi-gat: A graphical attention-based hierarchical multimodal
  representation learning approach for human activity recognition.
\newblock {\em IEEE Robotics and Automation Letters}, 6(2):1729--1736, 2021.

\bibitem{A18}
Petar Velickovic, Guillem Cucurull, Arantxa Casanova, Adriana Romero, Pietro
  Lio, and Yoshua Bengio.
\newblock Graph attention networks.
\newblock {\em stat}, 1050:20, 2017.

\bibitem{A19}
Boris Knyazev, Graham~W Taylor, and Mohamed Amer.
\newblock Understanding attention and generalization in graph neural networks.
\newblock {\em Advances in neural information processing systems}, 32, 2019.

\bibitem{A20}
Shaked Brody, Uri Alon, and Eran Yahav.
\newblock How attentive are graph attention networks?
\newblock {\em arXiv preprint arXiv:2105.14491}, 2021.

\bibitem{A21}
Xinyou Yin, JAN Goudriaan, Egbert~A Lantinga, JAN Vos, and Huub~J Spiertz.
\newblock A flexible sigmoid function of determinate growth.
\newblock {\em Annals of botany}, 91(3):361--371, 2003.

\bibitem{A22}
Gongde Guo, Hui Wang, David Bell, Yaxin Bi, and Kieran Greer.
\newblock Knn model-based approach in classification.
\newblock In {\em OTM Confederated International Conferences" On the Move to
  Meaningful Internet Systems"}, pages 986--996. Springer, 2003.

\end{thebibliography}

\end{document}